\pdfoutput=1
\documentclass[letterpaper, 10 pt, conference]{ieeeconf}

\IEEEoverridecommandlockouts

\usepackage{epsfig}
\usepackage{times}
\usepackage{amsmath}
\usepackage{amssymb}

\usepackage{times}
\usepackage{graphicx,stfloats}
\DeclareGraphicsExtensions{.pdf,.png,.jpg}
\DeclareMathOperator*{\argmax}{argmax}

\usepackage{makecell}
\usepackage{booktabs}
\usepackage{multirow}
\usepackage{multicol}
\usepackage{here}
\usepackage{soul, color}
\usepackage{caption}
\usepackage{anyfontsize} 
\PassOptionsToPackage{hyphens}{url}
\usepackage{hyperref}
\usepackage{booktabs}  
\usepackage{cuted}
\usepackage{capt-of}
\usepackage{etoolbox}
\usepackage{cleveref}
\usepackage{comment}
\usepackage{threeparttable}
\usepackage{tikz}

\usepackage[letterpaper,
            left=48pt, right=56pt,
            top=57pt,  bottom=36pt,
            includeheadfoot
           ]{geometry}

\title{\LARGE \bf
Training-Free Target-Aware Refinement of 3D Gaussian Language Map }

\author{%
Sibaek Lee$^{1}$, Kyeongsu Kang$^{1}$, Seongbo Ha$^{1}$, Yewon Jeon$^{1}$, Joonyeol Choi$^{1}$, Daegeol Ko$^{1}$, Jaehyuk Hur$^{1}$,\\
Taemin Choi$^{2}$, Minsu Kim$^{2}$, Hyungjoo Ahn$^{2}$, Wonsok Yoo$^{2}$, and Hyeonwoo Yu$^{1}$%
\thanks{This work has been submitted to the IEEE for possible publication. Copyright may be transferred without notice, after which this version may no longer be accessible.
$^{1}$Sungkyunkwan University, Suwon, Republic of Korea. {\tt\footnotesize \{lmjlss, thithin0821, sobo3607, yw8431, joonyeol99, kjs4753, jack1008, hwyu\}@g.skku.edu} $^{2}$Samsung Electronics, Republic of Korea. {\tt\footnotesize \{tm77.choi, minsuya.kim, hyungjoo.ahn, wonsok.yoo\}@samsung.com}}%
}

\begin{document}

\IEEEaftertitletext{%
\begin{center}
\vspace{-1.5\baselineskip}%
\includegraphics[width=\textwidth]{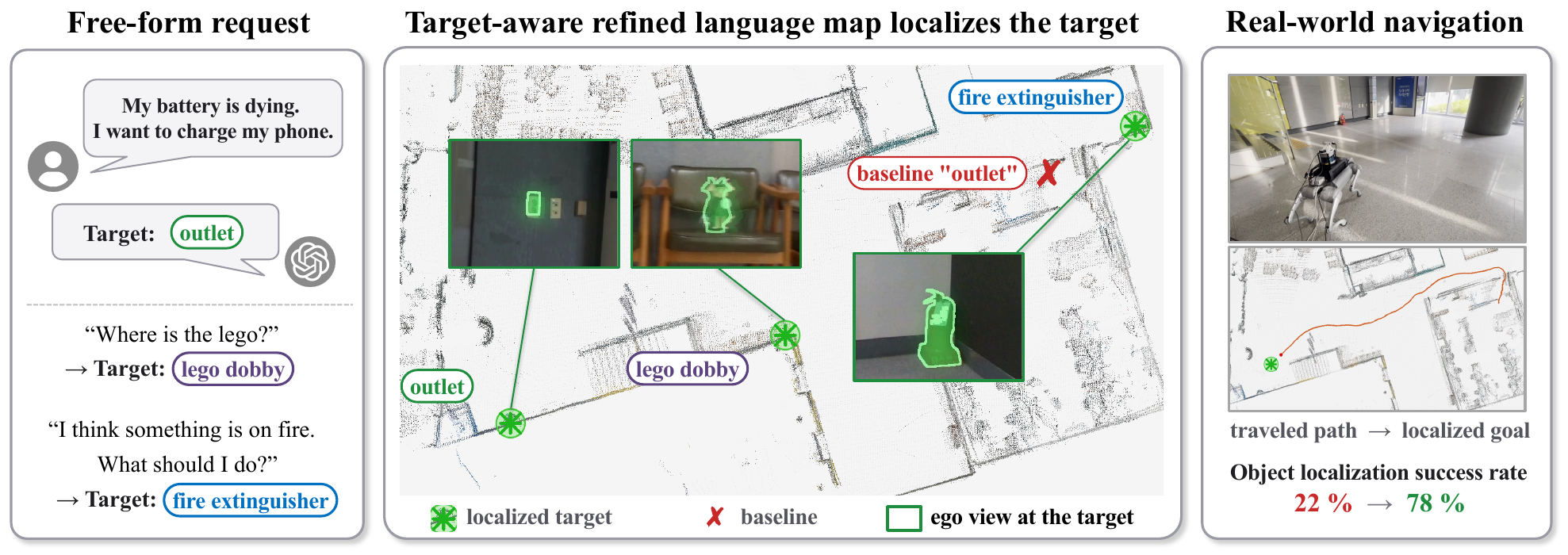}
\captionof{figure}{Training-free target-aware refinement of a 3D Gaussian language map. An LLM turns a free-form request into a target word, and the refined map localizes it (green stars), whereas the baseline misplaces the ``outlet'' (red cross). The localized goal then guides a quadruped robot.}
\label{fig:teaser}
\vspace{-0.5\baselineskip}%
\end{center}
}

\maketitle
\thispagestyle{empty}
\pagestyle{empty}

\begin{abstract}
Distilling vision-language features into 3D Gaussian Splatting yields a language map that can be queried with arbitrary text. In deployment, however, a small set of user-specified targets must be reliably located, which current distilled maps often miss, while open-vocabulary queryability for other words must remain intact. We propose a training-free, target-aware refinement of an already-distilled language map. Guided by a text-promptable segmenter, the visual embedding inside each detected target region is blended with the target text embedding, a convex combination whose coefficient sets the amount of injected prior, while non-target regions are left untouched. Injecting too much overwrites appearance cues and general queryability, so we seek, for each target, the smallest coefficient that still lets it win, and we obtain every coefficient in closed form, leaving no hyperparameters to tune. We further identify a failure mode of injection in language-goal localization. Text embeddings share a common component, and injecting them makes the standard cosine read-out query-agnostic, which a centering step in the read-out removes. Across three benchmarks, the refinement consistently improves open-vocabulary segmentation, roughly doubling the base mIoU on Replica, and localizes language goals more reliably than the base map. It also transfers as a plug-in to off-the-shelf field-based and object-level language maps and supports a real-world demonstration in which a legged robot navigates to natural-language goals.
Project page and code: \url{https://lab-of-ai-and-robotics.github.io/target-aware-langmap/}
\end{abstract}

\section{INTRODUCTION}

A reliable spatial representation is central to robotic autonomy, since a robot must understand its environment before it can act within it across tasks such as localization, navigation, and manipulation. The recent emergence of Neural Radiance Fields (NeRF)~\cite{mildenhall2021nerf} and 3D Gaussian Splatting (3DGS)~\cite{kerbl20233d} has made photorealistic spatial representation possible. Moreover, by combining these representations with vision language models~\cite{radford2021learning}, a scene is no longer restricted to a fixed set of semantic labels, and open-vocabulary NeRF and 3DGS representations~\cite{kerr2023lerf,zhou2024feature,qin2024langsplat} have become possible, allowing the space to be queried with arbitrary natural language.

However, existing methods for building such language maps tend to fall toward one of two ends of a spectrum, and we instead pursue the practical middle ground, precise localization of user-specified targets without giving up open-vocabulary generality. Closed-set methods are trained per scene with semantic supervision~\cite{li2024sgs,zhu2024sni} and are accurate on their predefined labels, but cannot be queried with any word outside that set. Open-vocabulary methods instead distill vision-language features and respond to arbitrary text~\cite{kerr2023lerf,qin2024langsplat,zhou2024feature}. Yet because the distillation blends surrounding context into each feature, the resulting relevancy can be spatially diffuse, so a query often yields a plausible heatmap whose peak is not always precise enough to serve directly as a navigation goal. We first tried to sharpen such maps with a training-free post-processing, but found that the limitation is already present in the 2D features themselves, whose top-ranked retrieval is often wrong before any lifting to 3D. Purely post-hoc processing is therefore insufficient, which motivates injecting the target text directly into the features.

To this end, we propose a training-free, target-aware refinement of an already-distilled language map. Given a small set of user-specified target words, we localize each target in every training view with a text-promptable segmenter, retain the open-vocabulary base from a class-agnostic segmenter, and encode both with a vision-language model. Within each target region the visual embedding is then blended with the target text embedding, a convex combination whose coefficient decides how much prior is injected, while non-target regions are left untouched. Such coefficients are conventionally treated as hyperparameters, swept over a grid or set by heuristics. Injecting too much overwrites the region's visual embedding and erodes the appearance cues and general queryability that make the map open-vocabulary. We therefore formulate the problem as finding the smallest coefficient that still lets each target win, and we solve it in closed form. Each detection yields this minimum as a flip point, and the per-target strength is read off the knee of the capture curve that the flip points form, leaving no hyperparameters to tune. This minimal injection, together with the untouched non-target regions, is what keeps the refined map open-vocabulary.

Because the refinement acts only on the 2D features before distillation, it is pipeline-agnostic and drops into different language maps as a plug-in module, from dense feature fields to object-level maps. It also carries over to language-goal localization, where the map must turn a query word into a 3D goal position. We observe a failure mode rooted in the geometry of contrastive vision-language embeddings~\cite{liang2022mind,mu2017all}. Text embeddings cluster around a shared direction, so injecting them plants this common component in the map, and under the standard cosine read-out every query then retrieves the same dominant regions. A centering step in the read-out removes this common mode. Notably, centering barely changes the base map but is decisive on the refined one, after which the refined map localizes goals more reliably than the base across Replica~\cite{straub2019replica}, ScanNet~\cite{dai2017scannet}, and HM3D-Semantics~\cite{yadav2023habitat}. We summarize our contributions as follows:
\begin{itemize}
\item A training-free, target-aware refinement that makes user-specified targets reliably retrievable while keeping the map open-vocabulary, with each target's injection strength obtained in closed form instead of tuned.
\item A centered read-out that identifies and removes the query-agnostic bias of injection, turning the refined map into a reliable source of language-goal locations.
\item Consistent gains in segmentation and localization on three benchmarks, plug-in transfer to field-based and object-level maps, and a real-world demonstration on a legged robot.
\end{itemize}

\section{Related work}
\subsection{Open-vocabulary 3D language maps}
Semantic mapping has long attached labels to reconstructed geometry, but classical systems remain closed set, recognizing only the categories their segmenters were trained on~\cite{mccormac2017semanticfusion}. Open-vocabulary maps remove this limit by distilling vision-language embeddings into 3D, first over point clouds and implicit fields~\cite{peng2023openscene,jatavallabhula2023conceptfusion,kerr2023lerf} and more recently into 3D Gaussian Splatting~\cite{qin2024langsplat,zhou2024feature,wu2024opengaussian}, with closed-form solvers reducing the distillation cost~\cite{xiong2025splat,jun2025dr} and online variants embedding language during SLAM~\cite{lee2025lego,ha2026langgs}. Dedicated benchmarks now track the open-vocabulary quality of such maps~\cite{kassab2026openlex3d}. All of these works concern how to build the map, and to our knowledge none refines the language field of an already-built one, so our training-free, target-aware refinement is complementary to all of them.

\subsection{Language-goal localization and navigation}
A language map supports many downstream tasks, and this work concerns the one closest to deployment, turning a query word into a navigation goal. Spatial memories built from vision-language features have enabled language-driven navigation over grids and implicit fields~\cite{huang2023visual,shafiullah2022clip,lee2025lamp}, zero-shot object search~\cite{gadre2023cows}, scene graphs queried through language model reasoning~\cite{werby2024hierarchical,gu2024conceptgraphs}, and Gaussian memories for goal retrieval~\cite{zhou2025lagmemo}, evaluated under object-goal navigation protocols~\cite{batra2020objectnav,khanna2024goat}. Across these systems the goal is typically read out by raw cosine similarity between the query embedding and the stored features. We show that this standard read-out breaks once text embeddings are injected into the map, and that a centering step in the read-out restores and then surpasses the base retrieval.

\section{Method}
\label{sec:method}

\begin{figure*}[t]
\centering
\includegraphics[width=\textwidth]{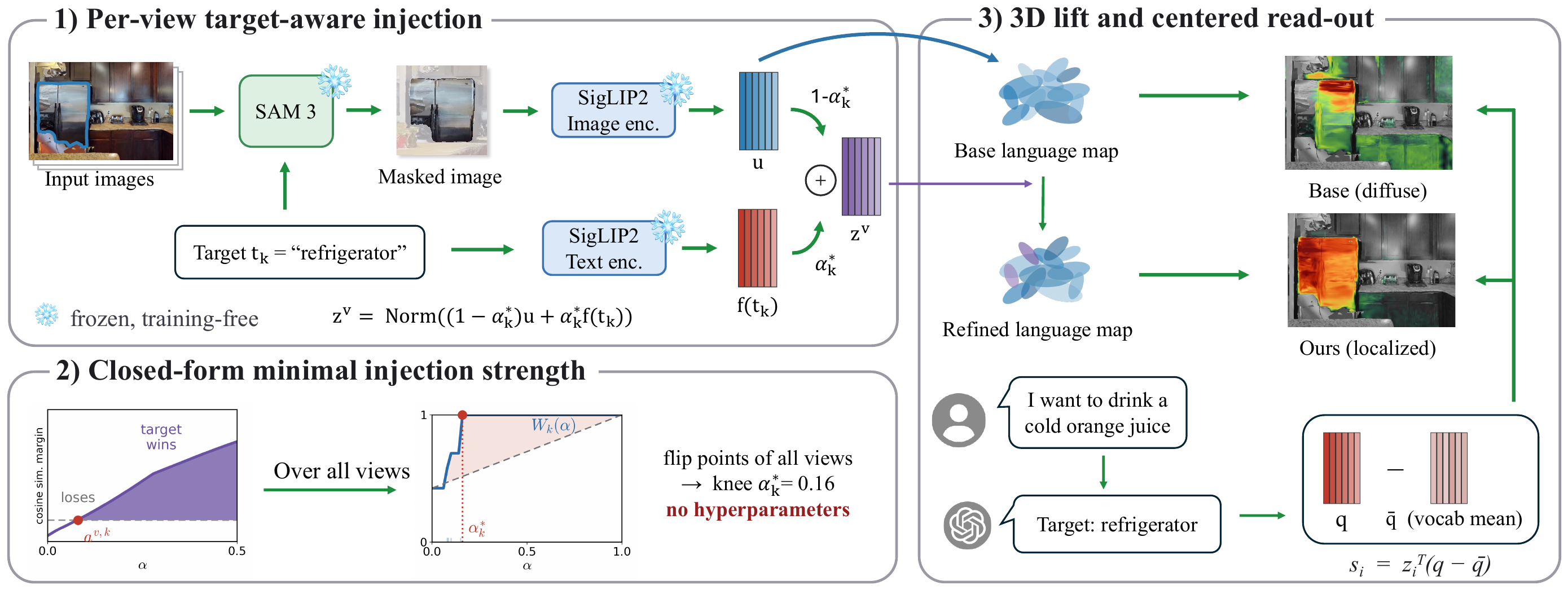}
\caption{Overview. 1) SAM\,3 localizes each user-specified target in every training view, and the region embedding $u$ is blended toward the text anchor $f(t_k)$ with the per-target strength $\alpha_k^{*}$. 2) The flip point of each detection, the smallest $\alpha$ that wins, has a closed form, and the knee of the capture curve over all views sets $\alpha_k^{*}$ without hyperparameters. 3) The refined features are lifted onto the frozen geometry, and a centered read-out turns the map into 3D goal locations.}
\label{fig:method}
\end{figure*}

Our method refines an existing 3D Gaussian language map so that user-specified targets become reliably retrievable while the map stays open-vocabulary. For every training view we localize each target, blend its visual embedding toward the target text by a coefficient obtained in closed form, and lift the refined per-view features into 3D. At query time a centered read-out turns the refined map into 3D goal locations for navigation. Fig.~\ref{fig:method} summarizes the pipeline.

\subsection{Problem Setup}
\label{sec:setup}
We are given a posed 3DGS scene whose language field is built from multi-view vision-language features. Each Gaussian $i$ carries an L2-normalized feature $z_i \in \mathbb{R}^{D}$ in the joint embedding space, and a text query $t$ scores Gaussians by the cosine similarity $z_i^{\top} f(t)$, where $f(\cdot)$ denotes the text encoder. The application additionally provides a small set of target queries $\{t_k\}_{k=1}^{n}$ with text anchors $f(t_k)$. A target region is retrieved only when its features score higher for $t_k$ than for every competing anchor. Densely distilled features frequently fail this test, since distillation mixes surrounding context into each feature and the argmax drifts toward larger context categories.

In this setting, our goal is to refine the features so that two conditions hold. First, each target $t_k$ must win this comparison reliably, so that querying $t_k$ retrieves its own instances rather than the surrounding context that currently absorbs them. Second, the map must remain open-vocabulary, so features that belong to no target must stay unchanged, and even the target features should move only as much as needed. To this end the refinement introduces, for each target, one injection coefficient $\alpha_k \in [0,1]$ that blends the target features toward the text anchor as a convex combination of the two. The problem we solve is choosing each $\alpha_k$ as the smallest value for which $t_k$ still wins, with no tuned hyperparameter.

\subsection{Target-Aware Feature Injection}
\label{sec:inject}
We construct a target-aware feature map for every training view and then lift the per-view maps into 3D, in our implementation with a closed-form feature splatting solver~\cite{xiong2025splat}. For each view, a class-agnostic segmenter, instantiated with SAM\,2~\cite{ravi2025sam} in automatic mask generation mode, produces masks that cover the view, and a vision-language model, for which we use SigLIP\,2~\cite{tschannen2025siglip}, encodes each masked region into a $D$-dimensional embedding. These region embeddings form the open-vocabulary base of the map. For each target word $t_k$, a text-promptable segmenter, SAM\,3~\cite{carion2025sam}, localizes the target with a per-target mask in the same view. The shared encoder embeds that region into $u^{v,k}$, and the text anchor $f(t_k)$ comes from the paired text encoder. Regions outside the target masks keep their visual embeddings, so the base is preserved by construction. Targets that overlap each other are resolved smaller mask first, which is deterministic and favors the more specific detection.

Within each target region we form the convex combination
\begin{equation}
z^{v,k}(\alpha) \;=\; \mathrm{Norm}\!\big((1-\alpha)\,u^{v,k} + \alpha\, f(t_k)\big),
\label{eq:blend}
\end{equation}
where $\alpha=0$ keeps the pure visual embedding and $\alpha=1$ is a hard text paste. What remains is the choice of this coefficient, which we settle next in closed form.

\subsection{Closed-Form Minimal Injection}
\label{sec:knee}
For each view, the smallest coefficient at which the detected target region starts to win has a closed form. Write $u=u^{v,k}$, $\kappa=u^{\!\top}f(t_k)$, $\beta_j=u^{\!\top}f(t_j)$, and $\gamma_j=f(t_k)^{\!\top}f(t_j)$ for unit vectors. The winning condition $\cos(z^{v,k}(\alpha),f(t_k))\ge\cos(z^{v,k}(\alpha),f(t_j))$ shares the denominator $\lVert(1-\alpha)u+\alpha f(t_k)\rVert$ on both sides, so it reduces to the linear condition $(1-\alpha)\kappa+\alpha \ge (1-\alpha)\beta_j+\alpha\gamma_j$, whose boundary $a_j=\frac{\beta_j-\kappa}{\beta_j-\kappa+1-\gamma_j}$, clipped to $[0,1]$, is the point where the blend overtakes competitor $j$. The view's flip point is the coefficient that beats all competitors,
\begin{equation}
a^{v,k} \;=\; \max_j a_j \;=\; \min\{\alpha\in[0,1] : z^{v,k}(\alpha)\ \text{retrieves}\ t_k\},
\label{eq:flip}
\end{equation}
which is $0$ when the region already wins and $1$ when it never does. Computing it costs a handful of dot products per view and introduces no extra constants to tune.

\begin{figure}[t]
\centering
\includegraphics[width=\linewidth]{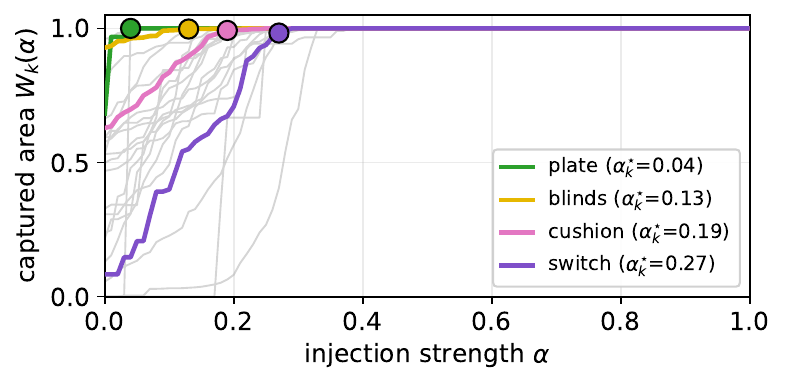}
\caption{Capture curves and knees on Replica room0. Each curve is a target's capture curve $W_k(\alpha)$ and each dot marks its knee $\alpha_k^{*}$, with the remaining targets in grey. Targets whose views flip late, such as switch, receive a larger coefficient than targets the base almost retrieves, such as plate.}
\label{fig:capture}
\end{figure}

A single per-target coefficient $\alpha_k$ cannot be minimal for every view at once, since each view flips at its own point, and injecting each view exactly at its own flip point would win by only a zero margin while varying the strength across the views that are later averaged. We therefore aggregate the flip points into an area-weighted capture curve,
\begin{equation}
W_k(\alpha) \;=\; \frac{\sum_{v} A^{v,k}\,\mathbb{1}[a^{v,k}\le\alpha]}{\sum_{v} A^{v,k}},
\label{eq:capture}
\end{equation}
the non-decreasing fraction of the target's detected area that wins at strength $\alpha$, where $A^{v,k}$ denotes mask area. The per-target strength is read off this curve at its knee, the point of maximum height above the chord joining the endpoints,
\begin{equation}
\alpha_k^{*} \;=\; \argmax_{\alpha\in[0,1]}\big(W_k(\alpha) - \mathrm{chord}_k(\alpha)\big).
\label{eq:alpha_star}
\end{equation}
The knee is where additional injection stops buying capture, a scale-invariant operating point that assumes no exchange rate between gain and cost. It may leave the most stubborn views unrecovered, trading complete capture for minimal overwrite, and after multi-view averaging the captured majority typically suffices for map-level retrieval. Fig.~\ref{fig:capture} shows the resulting curves and knees for one scene. The knee keeps $\alpha_k^{*}$ small and adaptive, with mean strengths around $0.1$ to $0.2$ in our experiments and exactly $0$ for targets the base already retrieves, so already-correct features are left untouched.

The per-view maps, SAM\,2 regions carrying their visual embeddings and SAM\,3 regions carrying $z^{v,k}(\alpha_k^{*})$, are then lifted by the solver, which multi-view averages the injections. Because every view is nudged only slightly and the views are averaged, the target wins class-level retrieval while its instance-level appearance survives, whereas a hard paste collapses it (Sec.~\ref{sec:alpha_analysis}). We decide the coefficient per view before aggregation, since fixing a single coefficient against the averaged map-level feature would lose the per-view adaptivity that recovers the hardest targets.

\subsection{Centered Localization Read-out}
\label{sec:readout}
The refined map is $N$ Gaussians with positions $x_i$ and unit-norm features $z_i$. What remains is the read-out, the procedure that turns a text query $t$ with anchor $q=f(t)$ into a few 3D goal locations. The natural choice, ranking Gaussians by the raw cosine $z_i^{\top}q$, works on the base map but fails once the map is refined. Injection pulls the target features toward the text anchors, which all lie in a narrow cone of the embedding space~\cite{liang2022mind,mu2017all}, so every injected feature picks up a large shared component, a common mode, pointing into that cone. Because this common mode is the same for any query, it dominates the score, and different query words end up ranking the Gaussians in nearly the same order.

We remove it by centering the query with the mean anchor $\bar q=\tfrac{1}{|\mathcal V|}\sum_{w\in\mathcal V}f(w)$ over a query vocabulary $\mathcal V$,
\begin{equation}
s_i=z_i^\top\!\big(q-\bar q\big)
   =\underbrace{z_i^\top q}_{\text{query response}}-\underbrace{z_i^\top\bar q}_{\text{common mode}},
\label{eq:centering}
\end{equation}
which equivalently subtracts from each Gaussian its mean similarity over $\mathcal V$, leaving only its query-specific response. The segmentation argmax is unaffected, since the subtracted term is shared by every class anchor at a given Gaussian, so centering matters only for the per-query ranking over Gaussians. Subtracting this shared component barely changes the base map but becomes decisive here, since injection is what plants the common mode in the first place. The step is feed-forward, leaves the map itself unchanged, and needs only a generic vocabulary, as a scene-independent word list works equally well.

To extract goals we keep the top-$p\%$ Gaussians by $s_i$, cluster their positions with DBSCAN, and rank the clusters by their similarity mass $\sum_{i\in c}\max(s_i,0)^2$. The top-$K$ clusters return their similarity-weighted centroids
\begin{equation}
g_c=\Big(\textstyle\sum_{i\in c}\max(s_i,0)^2\,x_i\Big)\Big/\textstyle\sum_{i\in c}\max(s_i,0)^2
\label{eq:candidate}
\end{equation}
as the goal candidates. The extraction constants, the kept fraction $p$, the cluster radius, and $K$, are fixed once across all scenes and datasets, so the pipeline stays free of per-scene tuning. This centered read-out turns the segmentation-oriented refinement into a reliable localizer for navigation (Sec.~\ref{sec:localization}).

\section{Experiments}

\subsection{Experimental Setup}
\subsubsection{Datasets and metrics}
We evaluate on 8 Replica~\cite{straub2019replica}, 10 ScanNet~\cite{dai2017scannet}, and the 4 multi-room HM3D-Semantics~\cite{yadav2023habitat} scenes that LagMemo~\cite{zhou2025lagmemo} curates from GOAT-Bench~\cite{khanna2024goat}. On each per-scene map we run two evaluations, rendered-view open-vocabulary segmentation and language-goal localization for navigation. For segmentation the target queries are each scene's annotated categories, which serve as both the deployment vocabulary and the evaluation set. For each test view we label every pixel by the $\argmax$ cosine similarity to the GT class names and score mIoU and mAcc against the rendered GT, ignoring unlabeled pixels. On HM3D-Sem, whose scenes carry over a thousand free-form categories, we follow the common practice of restricting to frequent categories~\cite{rozenberszki2022language} and keep per scene those covering at least $0.5\%$ of the labeled pixels. The navigation-oriented localization study of Sec.~\ref{sec:localization} instead queries the shared object-level HOV-SG vocabulary~\cite{werby2024hierarchical}. Beyond the main comparison we run two further studies, each with a single protocol across all scenes. Table~\ref{tab:alpha_extra} ablates injection strength, comparing our adaptive knee against the base and a range of fixed strengths. To check that injection preserves open-vocabulary behavior, the held-out study in Table~\ref{tab:heldout} splits each scene's classes 50/50 into an injected target set and a never-injected held-out set, scores both with the full class set, and reports the held-out change from the base.

\subsubsection{Implementation details}
Our pipeline builds a base from SAM\,2~\cite{ravi2025sam} masks encoded by SigLIP\,2~\cite{tschannen2025siglip} and applies the proposed hyperparameter-free refinement to it, lifting the per-view features with the feature-splatting solver~\cite{xiong2025splat}. On Replica and ScanNet the geometry is reconstructed from RGB alone, matching the RGB-only baselines, while the house-scale HM3D-Sem scenes, where RGB-only reconstruction is unreliable, are initialized with depth following the RGB-D setting of LagMemo~\cite{zhou2025lagmemo}. The language distillation and refinement are identical in both the RGB and RGB-D settings. On HM3D-Sem the geometry is trained on every rendered frame while the language stages of feature extraction, injection, and evaluation run on every fifth frame. The read-out constants are fixed across all scenes and datasets, a kept fraction of $4\%$ for the centered read-out and $0.5\%$ for the raw read-out, so each is scored at its own operating point, with a cluster radius of $0.3$\,m and $K{=}5$.

\subsubsection{Baselines}
We compare against two field-based 3DGS language maps listed in Table~\ref{tab:seg_main}, each run in its published configuration and scored with our protocol. LangSplat~\cite{qin2024langsplat} distills three-granularity SAM regions encoded by OpenCLIP~\cite{ilharco2021openclip} into a per-scene autoencoder latent, and Feature-3DGS~\cite{zhou2024feature} distills LSeg~\cite{li2022language} features into the Gaussians. For each baseline, +Ours applies the same refinement to its per-view features, changing only the text anchors to that baseline's paired text encoder. The Base row is our SAM\,2 + SigLIP\,2 map without refinement, the reference practice of existing open-vocabulary maps. For localization in Table~\ref{tab:local} the baseline is this same base map queried with the standard cosine read-out, against which we isolate the effect of the refinement and of the centering step. Table~\ref{tab:local_ext} additionally runs two object-level systems, HOV-SG~\cite{werby2024hierarchical} and ConceptGraphs~\cite{gu2024conceptgraphs}, end to end in their published configurations, with + Ours applying the same refinement to their per-frame region features via their paired CLIP encoder.

\begin{table}[t]
\centering
\setlength{\tabcolsep}{3.5pt}
\footnotesize
\caption{Open-vocabulary segmentation, zero-shot rendered-view mIoU and mAcc, dataset means. Bold marks the best per column.}
\label{tab:seg_main}
\begin{tabular}{lcccccc}
\toprule
 & \multicolumn{2}{c}{Replica} & \multicolumn{2}{c}{ScanNet} & \multicolumn{2}{c}{HM3D-Sem} \\
\cmidrule(lr){2-3}\cmidrule(lr){4-5}\cmidrule(lr){6-7}
Method & mIoU & mAcc & mIoU & mAcc & mIoU & mAcc \\
\midrule
LangSplat~\cite{qin2024langsplat}   & 0.176 & 0.280 & 0.164 & 0.297 & 0.049 & 0.079 \\
LangSplat + Ours                    & 0.315 & 0.416 & 0.366 & 0.518 & 0.104 & 0.185 \\
\midrule
Feature-3DGS~\cite{zhou2024feature} & 0.263 & 0.345 & 0.571 & 0.708 & 0.127 & 0.179 \\
Feature-3DGS + Ours                 & 0.324 & 0.406 & \textbf{0.611} & \textbf{0.750} & 0.170 & 0.238 \\
\midrule
Base (SAM2+SigLIP2) & 0.207 & 0.311 & 0.355 & 0.512 & 0.144 & 0.199 \\
\textbf{Ours}                       & \textbf{0.436} & \textbf{0.546} & 0.542 & 0.694 & \textbf{0.231} & \textbf{0.313} \\
\bottomrule
\end{tabular}
\end{table}

\begin{table}[t]
\centering
\setlength{\tabcolsep}{3pt}
\footnotesize
\caption{Injection-strength ablation on all three benchmarks. The last row gives $\bar\alpha^{*}$ as Replica/ScanNet/HM3D.}
\label{tab:alpha_extra}
\begin{tabular}{lcccccc}
\toprule
 & Base & $\alpha{=}0.1$ & $\alpha{=}0.2$ & Ours & $\alpha{=}0.5$ & $\alpha{=}1$ \\
\midrule
\multicolumn{7}{l}{\textit{Replica}} \\
\quad mIoU & 0.207 & 0.360 & 0.416 & 0.436 & 0.461 & 0.471 \\
\quad mAcc & 0.311 & 0.462 & 0.525 & 0.546 & 0.576 & 0.588 \\
\multicolumn{7}{l}{\textit{ScanNet}} \\
\quad mIoU & 0.355 & 0.501 & 0.541 & 0.542 & 0.557 & 0.560 \\
\quad mAcc & 0.512 & 0.659 & 0.697 & 0.694 & 0.711 & 0.713 \\
\multicolumn{7}{l}{\textit{HM3D}} \\
\quad mIoU & 0.144 & 0.190 & 0.215 & 0.231 & 0.258 & 0.262 \\
\quad mAcc & 0.199 & 0.266 & 0.294 & 0.313 & 0.350 & 0.365 \\
\midrule
$\bar\alpha^{*}$ & 0.00 & 0.10 & 0.20 & 0.17/0.12/0.14 & 0.50 & 1.00 \\
\bottomrule
\end{tabular}
\vspace{6pt}

\includegraphics[width=\linewidth]{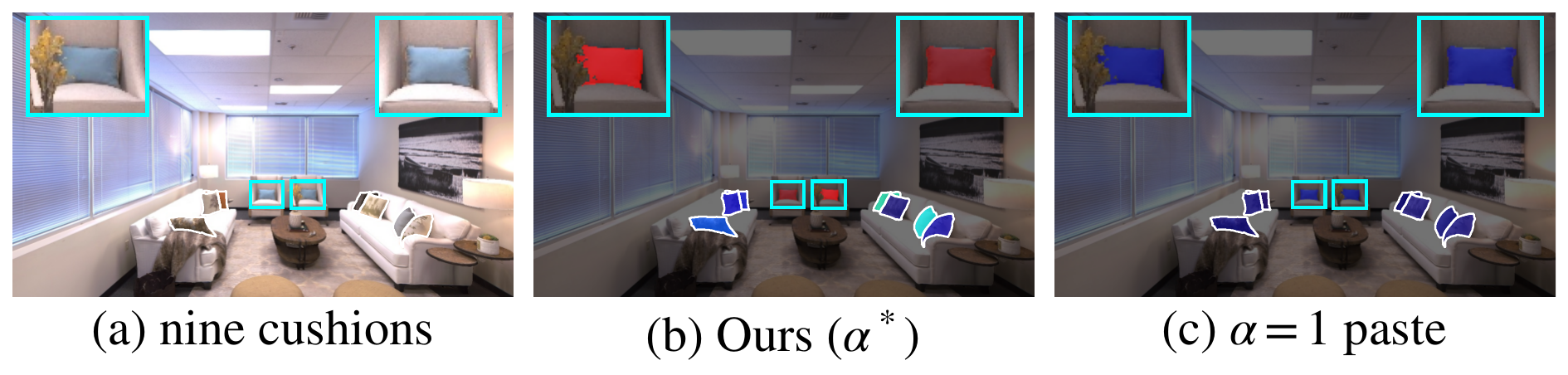}
\captionof{figure}{Instance-level discrimination on room0 cushions, colored by the differential relevancy for ``blue cushion''. The GT blue pair is boxed in cyan and magnified, and the remaining cushions are outlined in white. Ours preserves the distinction that the $\alpha{=}1$ paste collapses.}
\label{fig:instattr}
\end{table}

\subsection{Open-vocabulary segmentation}
\label{sec:analysis}
Our initial attempt, a training-free entropy-based post-processing, degraded mIoU on two language maps. The failure originates in the 2D features themselves. Even where a target region is segmented almost perfectly, the text-cosine ranks the correct word far down the list, so the region does not win its own query. A global post-hoc reweighting cannot fix this, which is why we inject the target text directly, small and selective, to preserve the original discriminability.

Our target-aware refinement instead raises the SAM\,2+SigLIP\,2 base from $0.207$ to $0.436$ mIoU on Replica. Most of the gain comes from targets the base misses entirely, such as picture, vase, and rug, which the refinement recovers from near-zero IoU to well above $0.4$, while already-retrieved classes are preserved. The injection stays small and adaptive, with a mean strength of $\bar\alpha^{*}=0.17$. Classes never detected by SAM\,3 remain at zero, so the residual bottleneck there is the text-promptable segmenter's recall rather than the injection mechanism.

The refinement also works as a plug-in on off-the-shelf maps, where the only change is that the text anchors come from each map's own text encoder. On LangSplat it raises mIoU by $79\%$ to $123\%$ across the three datasets. On Feature-3DGS it lifts mIoU by $23\%$ on Replica and $34\%$ on HM3D-Sem, improving all twelve of their scenes, and by a more modest $4.0$ points on ScanNet. The injected strength adapts to the base, shrinking where the base is already strong, with a correlation of $-0.80$ between base mIoU and mean strength across all scenes of the three datasets, so the gains concentrate where the base is weak.

\begin{table}[t]
\centering
\setlength{\tabcolsep}{4pt}
\footnotesize
\caption{Held-out generalization in mAcc, with each scene's classes split evenly at random into an injected target half and a held-out half. $\Delta$ columns give the relative change from the base in \%, and Average is the three-dataset mean.}
\label{tab:heldout}
\begingroup
\scriptsize
\setlength{\tabcolsep}{2.5pt}
\begin{tabular}{lcccccccc}
\toprule
 & \multicolumn{4}{c}{\textit{Replica}} & \multicolumn{4}{c}{\textit{ScanNet}} \\
\cmidrule(lr){2-5}\cmidrule(lr){6-9}
$\alpha$ & tgt & held & $\Delta$tgt & $\Delta$held & tgt & held & $\Delta$tgt & $\Delta$held \\
\midrule
Base ($\alpha{=}0$) & 0.271 & 0.349 & --- & --- & 0.448 & 0.578 & --- & --- \\
Fixed $0.1$         & 0.409 & 0.334 & $+50.9$  & $-4.3$  & 0.662 & 0.530 & $+47.8$ & $-8.3$ \\
Fixed $0.2$         & 0.492 & 0.315 & $+81.5$  & $-9.7$  & 0.736 & 0.511 & $+64.3$ & $-11.6$ \\
Fixed $0.5$         & 0.558 & 0.268 & $+105.9$ & $-23.2$ & 0.775 & 0.491 & $+73.0$ & $-15.1$ \\
Fixed $1.0$         & 0.582 & 0.257 & $+114.8$ & $-26.4$ & 0.782 & 0.484 & $+74.6$ & $-16.3$ \\
Ours (adaptive)     & 0.498 & 0.320 & $+83.8$  & $-8.3$  & 0.676 & 0.530 & $+50.9$ & $-8.3$ \\
\midrule
 & \multicolumn{4}{c}{\textit{HM3D-Sem}} & \multicolumn{4}{c}{\textit{Average}} \\
\cmidrule(lr){2-5}\cmidrule(lr){6-9}
Base ($\alpha{=}0$) & 0.168 & 0.226 & --- & --- & 0.296 & 0.384 & --- & --- \\
Fixed $0.1$         & 0.268 & 0.206 & $+59.5$  & $-8.8$  & 0.446 & 0.357 & $+51.0$  & $-7.2$ \\
Fixed $0.2$         & 0.310 & 0.200 & $+84.5$  & $-11.5$ & 0.513 & 0.342 & $+73.4$  & $-11.0$ \\
Fixed $0.5$         & 0.392 & 0.176 & $+133.3$ & $-22.1$ & 0.575 & 0.312 & $+94.5$  & $-18.9$ \\
Fixed $1.0$         & 0.418 & 0.163 & $+148.8$ & $-27.9$ & 0.594 & 0.301 & $+100.9$ & $-21.6$ \\
Ours (adaptive)     & 0.335 & 0.195 & $+99.4$  & $-13.7$ & 0.503 & 0.348 & $+70.1$  & $-9.4$ \\
\bottomrule
\end{tabular}
\endgroup
\end{table}

\subsection{Why minimal injection}
\label{sec:alpha_analysis}

A natural question is why we do not paste the target text directly at $\alpha{=}1$ or pick one fixed $\alpha$ for every target. On the closed-set metric, mIoU rises monotonically with injection strength and the $\alpha{=}1$ hard paste is highest (Table~\ref{tab:alpha_extra}). Our adaptive refinement reaches $0.436$ on Replica, higher than any fixed $\alpha$ of comparable mean strength and within $8\%$ of the hard paste. But this metric can be gamed, since it rewards pasting the very word it scores against, so a larger $\alpha$ is favored regardless of whether the region still carries any appearance.

The cost of hard paste appears as soon as a query needs appearance. Replica room0 contains nine cushions of different colors. Under $\alpha{=}1$ their per-instance features collapse to essentially one vector, with mean pairwise cosine $0.999$, so ``blue cushion'' cannot prefer the blue ones in principle. Under the differential relevancy $\cos(z,f(q))-\cos(z,f(\text{class}))$, which cancels the class component, the hard paste ranks neither blue cushion first, whereas our adaptive map preserves base-level appearance diversity, pairwise cosine $0.883$ vs base $0.873$, and ranks the two blue cushions top-2, matching the uninjected base (Fig.~\ref{fig:instattr}). In short, adaptive $\alpha^{*}$ retains most of the closed-set gain while keeping the visual information that hard paste burns.

A complementary held-out analysis makes the same point at the open-vocabulary level (Table~\ref{tab:heldout}), reported in mAcc since held-out mIoU is dominated by a few large background classes. Across all three datasets a larger fixed $\alpha$ trades held-out accuracy for target accuracy, degrading the held-out classes monotonically, by up to $26\%$ on Replica, $16\%$ on ScanNet, and $28\%$ on HM3D-Sem at $\alpha{=}1$. Our adaptive map instead keeps the held-out loss small, from $-8\%$ on Replica and ScanNet to $-14\%$ on the fine-grained HM3D-Sem set, while still capturing most of the target gain, a better trade-off than any fixed $\alpha$ of comparable target gain. Minimal, adaptive injection is what preserves open-vocabulary performance beyond the target set.

\subsection{Language-goal localization}
\label{sec:localization}

\begin{table}[t]
\centering
\setlength{\tabcolsep}{4pt}
\footnotesize
\caption{Instance-aware localization success rate (\%), top-$K$, on our field-based map. Base follows the raw-similarity read-out of existing maps~\cite{zhou2025lagmemo,werby2024hierarchical}. All rows share the same clustering read-out and differ only in the feature and the centering step.}
\label{tab:local}
\begingroup
\scriptsize
\setlength{\tabcolsep}{2.5pt}
\begin{tabular}{lcccccccccc}
\toprule
 & \multicolumn{5}{c}{\textit{Replica}} & \multicolumn{5}{c}{\textit{ScanNet}} \\
\cmidrule(lr){2-6}\cmidrule(lr){7-11}
Method & 1 & 2 & 3 & 4 & 5 & 1 & 2 & 3 & 4 & 5 \\
\midrule
Base                    & 44.6 & 51.8 & 60.7 & 73.2 & 73.2 & 72.9 & \textbf{90.0} & 91.4 & \textbf{92.9} & \textbf{94.3} \\
Base + centering        & 57.1 & 62.5 & 66.1 & 75.0 & 76.8 & 71.4 & 87.1 & 87.1 & 88.6 & 90.0 \\
Ours w/o centering      & 26.8 & 42.9 & 44.6 & 48.2 & 50.0 & 30.0 & 42.9 & 50.0 & 50.0 & 50.0 \\
\textbf{Ours}           & \textbf{76.8} & \textbf{80.4} & \textbf{83.9} & \textbf{85.7} & \textbf{89.3} & \textbf{75.7} & \textbf{90.0} & \textbf{92.9} & \textbf{92.9} & 92.9 \\
\midrule
 & \multicolumn{5}{c}{\textit{HM3D-Sem}} & \multicolumn{5}{c}{\textit{Aggregate}} \\
\cmidrule(lr){2-6}\cmidrule(lr){7-11}
Base                    & 24.5 & \textbf{42.9} & 44.9 & 46.9 & 53.1 & 50.3 & 64.6 & 68.6 & 73.7 & 76.0 \\
Base + centering        & 24.5 & 34.7 & \textbf{51.0} & 53.1 & 57.1 & 53.7 & 64.6 & 70.3 & 74.3 & 76.6 \\
Ours w/o centering      & 10.2 & 14.3 & 16.3 & 22.4 & 22.4 & 23.4 & 34.9 & 38.9 & 41.7 & 42.3 \\
\textbf{Ours}           & \textbf{32.7} & 40.8 & 46.9 & \textbf{55.1} & \textbf{63.3} & \textbf{64.0} & \textbf{73.1} & \textbf{77.1} & \textbf{80.0} & \textbf{83.4} \\
\bottomrule
\end{tabular}
\endgroup
\vspace{6pt}

\includegraphics[width=\linewidth]{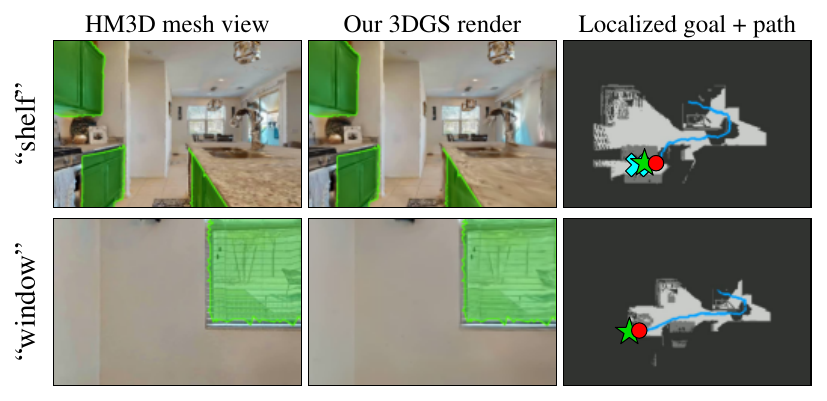}
\captionof{figure}{Language-goal navigation on HM3D-Sem. The green star marks the localized goal, the cyan circle the GT object, and the red trace shows the agent reaching it along the planned path. Left to right, first-person view, our 3DGS render, and the top-down map.}
\label{fig:nav}
\end{table}

\begin{table}[t]
\centering
\footnotesize
\caption{Refinement as a plug-in on object-level maps, instance-aware localization success rate (\%), top-$K$, same protocol as Table~\ref{tab:local}. HOV-SG~\cite{werby2024hierarchical} and ConceptGraphs~\cite{gu2024conceptgraphs} are queried through their own object-retrieval interfaces, and + Ours applies our refinement to their region features.}
\label{tab:local_ext}
\begingroup
\scriptsize
\setlength{\tabcolsep}{2.5pt}
\begin{tabular}{lcccccccccc}
\toprule
 & \multicolumn{5}{c}{\textit{Replica}} & \multicolumn{5}{c}{\textit{ScanNet}} \\
\cmidrule(lr){2-6}\cmidrule(lr){7-11}
Method & 1 & 2 & 3 & 4 & 5 & 1 & 2 & 3 & 4 & 5 \\
\midrule
HOV-SG~\cite{werby2024hierarchical}      & \textbf{14.3} & 16.1 & \textbf{26.8} & \textbf{26.8} & \textbf{30.4} & 78.6 & 82.9 & 87.1 & 95.7 & 98.6 \\
\;\;+ Ours                               & 12.5 & \textbf{19.6} & 21.4 & 21.4 & 25.0 & \textbf{90.0} & \textbf{90.0} & \textbf{92.9} & \textbf{98.6} & \textbf{100.0} \\
\midrule
ConceptGraphs~\cite{gu2024conceptgraphs} & \textbf{10.7} & \textbf{17.9} & \textbf{21.4} & \textbf{25.0} & \textbf{25.0} & 71.4 & 82.9 & 90.0 & 91.4 & 92.9 \\
\;\;+ Ours                               & 8.9 & 8.9 & 8.9 & 12.5 & 16.1 & \textbf{87.1} & \textbf{94.3} & \textbf{95.7} & \textbf{97.1} & \textbf{98.6} \\
\midrule
 & \multicolumn{5}{c}{\textit{HM3D-Sem}} & \multicolumn{5}{c}{\textit{Aggregate}} \\
\cmidrule(lr){2-6}\cmidrule(lr){7-11}
HOV-SG~\cite{werby2024hierarchical}      & 36.7 & 44.9 & 51.0 & \textbf{57.1} & \textbf{57.1} & 46.3 & 50.9 & 57.7 & \textbf{62.9} & \textbf{65.1} \\
\;\;+ Ours                               & \textbf{42.9} & \textbf{49.0} & \textbf{55.1} & \textbf{57.1} & \textbf{57.1} & \textbf{52.0} & \textbf{56.0} & \textbf{59.4} & 62.3 & 64.0 \\
\midrule
ConceptGraphs~\cite{gu2024conceptgraphs} & 20.4 & 26.5 & 30.6 & 36.7 & 40.8 & 37.7 & 46.3 & \textbf{51.4} & \textbf{54.9} & \textbf{56.6} \\
\;\;+ Ours                               & \textbf{30.6} & \textbf{30.6} & \textbf{36.7} & \textbf{42.9} & \textbf{42.9} & \textbf{46.3} & \textbf{49.1} & \textbf{51.4} & \textbf{54.9} & \textbf{56.6} \\
\bottomrule
\end{tabular}
\endgroup
\end{table}

We evaluate language-goal localization directly with an instance-aware success rate, reported at top-$K$ for $K{=}1\ldots5$. A predicted candidate counts as a hit if it lands within $1$\,m of any ground-truth instance of the queried class, following the official success radius of the Habitat ObjectNav challenge~\cite{batra2020objectnav}, also used by GOAT-Bench~\cite{khanna2024goat}. For the query vocabulary we adopt the $20$ object categories of the HOV-SG object-level localization benchmark~\cite{werby2024hierarchical}. Following HOV-SG, each scene is queried only with the categories present in its ground-truth annotation, with dataset-native synonyms mapped to the shared vocabulary\footnote{sofa$\to$couch, cushion$\to$pillow, indoor-plant$\to$plant, painting$\to$picture, pot$\to$flowerpot, tv-screen and television$\to$tv, books$\to$book, night stand$\to$nightstand, stair$\to$stairs, plus trivial spelling variants.}, which yields $56$ queries on Replica, $70$ on ScanNet, and $49$ on HM3D-Sem, $175$ in total. Candidates are extracted with the clustering read-out of Sec.~\ref{sec:readout}. The Base rows apply the same clustering to the raw cosine scores, retrieving goals by raw similarity as existing language maps do~\cite{huang2023visual,shafiullah2022clip,zhou2025lagmemo,werby2024hierarchical}, so the base stands in for the practice of these systems. Running such systems end to end would change the underlying map together with the read-out, whereas holding our map fixed isolates the effect of the refinement and of the centering from the representation itself. Since the goal coordinate comes directly from the map, the agent follows the shortest path to it, path efficiency is near-optimal by construction, and SPL reduces to the success rate, which we therefore report alone.

\begin{figure}[t]
\centering
\includegraphics[width=\linewidth]{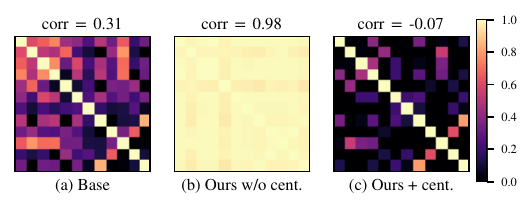}
\caption{Cross-query similarity correlation on a representative HM3D-Sem scene. Injection makes the refined map query-agnostic, and centering restores discriminability.}
\label{fig:centering}
\end{figure}

Table~\ref{tab:local} exposes a failure mode and its fix. Applying the refined map with the standard cosine read-out, Ours w/o centering, collapses far below the base on all three datasets, on ScanNet from $72.9$ to $30.0$ at top-1, because injection makes the map query-agnostic. As Fig.~\ref{fig:centering} shows, the per-Gaussian similarities of different query words become almost perfectly correlated, with the scene-averaged cross-query correlation growing from $0.28$ to $0.98$. Injection inflates a shared common mode that dwarfs the query-specific variation, raising the average similarity of every Gaussian to all queries from $0.056$ to $0.233$, so every query retrieves the same dominant injected object. The centering step of Sec.~\ref{sec:readout} removes this common mode. It helps the base only modestly, by $3.4$ points at top-1 over the $175$ queries, because the base has little common mode to remove, but on the refined map it is decisive.

Ours, the refined map with the centered read-out, beats the base at every $K$ on Replica and on the aggregate, $64.0$ vs $50.3$ at top-1, and beats the centering-matched base at every aggregate $K$ as well, by $10.3$ points at top-1, isolating a genuine contribution of the refinement itself rather than of centering alone. The gain is largest where the base is weakest, on Replica office1 from $17$ to $83$ at top-1, while on ScanNet, whose large furniture-level queries leave the base close to its ceiling, ours trails it by at most one query at every $K$. The ordering is unchanged at a looser $2$\,m radius, $76.0$ vs $62.9$ at top-1. The gain is also leak-free, staying identical when each query is excluded from its own centering mean and when a generic vocabulary unrelated to the scene is used instead.

Beyond this controlled comparison, Table~\ref{tab:local_ext} runs the object-level HOV-SG and ConceptGraphs end to end. Both are strong on the furniture-scale queries of ScanNet, where one clean feature per detected object avoids the context mixing of dense distillation, but collapse on the small, cluttered targets of Replica, where their mapping never instantiates the queried objects, so ours leads both by a wide margin on the aggregate, $64.0$ vs $46.3$ and $37.7$ at top-1. Injection transfers the refinement to this map family as well, raising the aggregate top-1 of HOV-SG to $52.0$ and of ConceptGraphs to $46.3$. The one block that does not improve is Replica, since the injection edits regions that already exist in the map and cannot create the missing small-object instances, which is precisely what the dense field representation provides.

An injection-strength sweep on HM3D-Sem confirms the mechanism. Without centering, top-1 success collapses from $24.5\%$ on the base to $10.2\%$ on the refined map, whereas with centering it rises from $24.5\%$ to $32.7\%$. The centered read-out even keeps improving past the knee, $38.8$ at a fixed $\alpha{=}1$ vs $32.7$, but that regime collapses the held-out performance of Table~\ref{tab:heldout} and the instance appearance of Fig.~\ref{fig:instattr}, so the knee remains the right operating point. Finally, Fig.~\ref{fig:nav} shows the read-out driving actual navigation, with the agent reaching each queried object along the planned path.

\subsection{Real-world demonstration}
\label{sec:realworld}
We deploy the full pipeline on a real building floor with a Unitree Go2 quadruped, shown in Fig.~\ref{fig:teaser}. A handheld rig sweeps the floor once, a LiDAR-inertial SLAM system~\cite{shan2020lio} estimates the sensor poses, and the posed RGB frames are lifted into a 3DGS language map by the same feature-splatting solver~\cite{xiong2025splat} and refined by our method, so the map construction is identical to the simulated setting. At run time the user issues a free-form natural-language request, a language-model front end selects the matching target word from the map's vocabulary, and the centered read-out turns that word into a 3D goal on the refined map. The robot localizes itself against the LiDAR map and plans toward the goal with A$^{*}$ on a 2D occupancy grid, so the language map supplies the goal and the LiDAR map supplies the pose and the grid in one shared frame. The refined map guides the robot to the correct object with a success rate of $78\%$, whereas the base map with the standard raw-similarity read-out reaches $22\%$, and the margin comes mostly from the more specific queries on which the base map fails.

\section{CONCLUSIONS}

We presented a training-free, target-aware refinement of 3D Gaussian language maps that makes user-specified targets reliably retrievable while keeping the map open-vocabulary. Each target's injection strength is obtained in closed form, so the refinement introduces no hyperparameters, and a centered read-out removes the query-agnostic common mode that injection plants, turning the refined map into a dependable source of language-goal locations. Across three benchmarks the same refinement improves both open-vocabulary segmentation and localization, and the gain carries over to a legged robot that navigates to natural-language goals on a real building floor. Because the refinement acts on per-view features before lifting, it drops into existing language-map pipelines, and extending it to online mapping and to targets observed only sparsely is a natural next step.

\section*{ACKNOWLEDGMENT}
This work was supported by Samsung Electronics.

{
\footnotesize
\bibliographystyle{IEEEtranS}
\bibliography{main}
}

\end{document}